\documentclass[10pt, conference]{IEEEtran}
\usepackage{graphicx} 
\usepackage{cite}
\usepackage{amsmath,amssymb,amsfonts}
\usepackage{algorithmic}
\usepackage{textcomp}
\usepackage{xcolor}
\usepackage[numbers]{natbib}
\usepackage{color, colortbl}
\usepackage{microtype}
\usepackage{subcaption}
\usepackage{booktabs} 

\usepackage{float}
\usepackage[colorlinks=true, allcolors=blue]{hyperref}
\usepackage{wrapfig}
\graphicspath{ {./images/} }
\usepackage{placeins}
\usepackage{dblfloatfix}

\usepackage[capitalize,noabbrev]{cleveref}

\def\BibTeX{{\rm B\kern-.05em{\sc i\kern-.025em b}\kern-.08em
    T\kern-.1667em\lower.7ex\hbox{E}\kern-.125emX}}

\begin{document}
\title{FedPAE: Peer-Adaptive Ensemble Learning for Asynchronous and Model-Heterogeneous Federated Learning}

\author{\IEEEauthorblockN{Brianna Mueller}
\IEEEauthorblockA{\textit{Department of Business Analytics} \\
\textit{University of Iowa}\\
Iowa City, USA \\
brianna-mueller@uiowa.edu}
\and
\IEEEauthorblockN{W. Nick Street}
\IEEEauthorblockA{\textit{Department of Business Analytics} \\
\textit{University of Iowa}\\
Iowa City, USA \\
nick-street@uiowa.edu}
\and
\IEEEauthorblockN{Stephen Baek}
\IEEEauthorblockA{\textit{School of Data Science} \\
\textit{University of Virginia}\\
Charlottesville, USA \\
mwn4yc@virginia.edu}
\and
\IEEEauthorblockN{Qihang Lin}
\IEEEauthorblockA{\textit{Department of Business Analytics} \\
\textit{University of Iowa}\\
Iowa City, USA \\
qihang-lin@uiowa.edu}
\and
\IEEEauthorblockN{Jingyi Yang}
\IEEEauthorblockA{\textit{Department of Information Systems} \\
\textit{New York University}\\
New York, USA \\
jy4057@stern.nyu.edu}
\and
\IEEEauthorblockN{Yankun Huang}
\IEEEauthorblockA{\textit{Department of Information Systems} \\
\textit{Arizona State University}\\
Tempe, USA \\
yankun.huang@asu.edu
}
}

\maketitle

\begin{abstract}
Federated learning (FL) enables multiple clients with distributed data sources to collaboratively train a shared model without compromising data privacy.  However, existing FL paradigms face challenges due to heterogeneity in client data distributions and system capabilities. Personalized federated learning (pFL) has been proposed to mitigate these problems, but often requires a shared model architecture and a central entity for parameter aggregation, resulting in scalability and communication issues. More recently, model-heterogeneous FL has gained attention due to its ability to support diverse client models, but existing methods are limited by their dependence on a centralized framework, synchronized training, and publicly available datasets. To address these limitations, we introduce Federated Peer-Adaptive Ensemble Learning (FedPAE), a fully decentralized pFL algorithm that supports model heterogeneity and asynchronous learning. Our approach utilizes a peer-to-peer model sharing mechanism and ensemble selection to achieve a more refined balance between local and global information. Experimental results show that FedPAE outperforms existing state-of-the-art pFL algorithms, effectively managing diverse client capabilities and demonstrating robustness against statistical heterogeneity. 
\end{abstract}


\section{Introduction}
\begin{figure}[!b]
\centering
\includegraphics[width=0.4\textwidth]{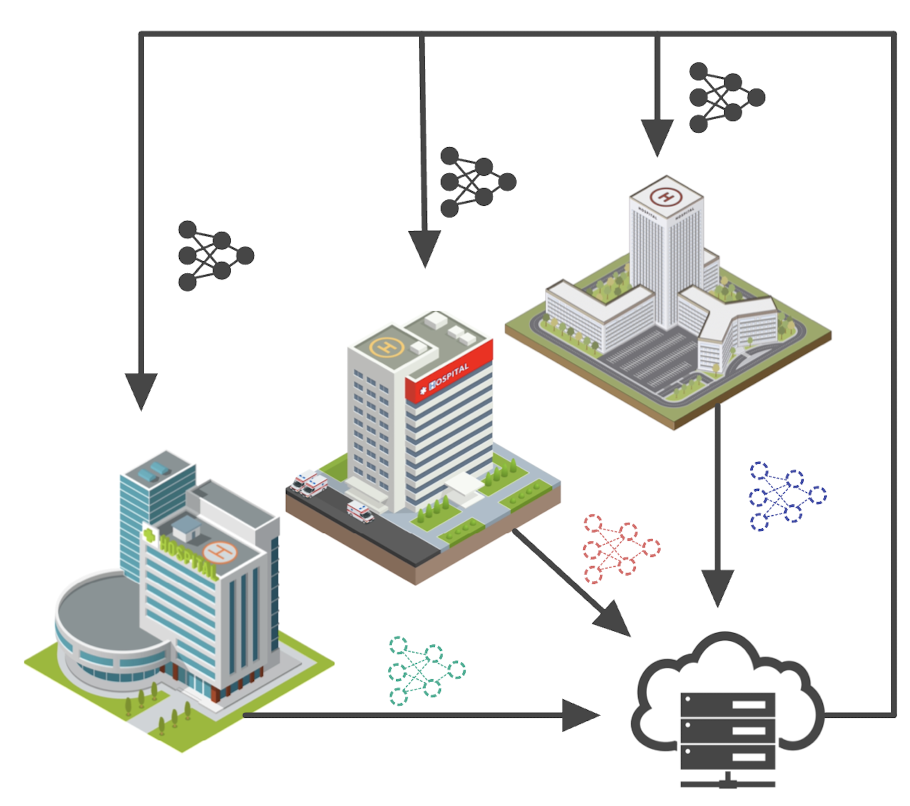}
\caption{A centralized and synchronous FL network}
\label{fig:federated}
\end{figure}

Federated learning has emerged as a promising approach for leveraging decentralized data sources for collaborative machine learning while preserving data privacy and security.  Traditionally, collaborative learning requires a central server to aggregate data from multiple sources for model training \cite{abdulrahman_survey_2021}. However, industries with privacy-sensitive data (e.g., healthcare, cybersecurity, finance) often face stringent regulations that inhibit centralized data aggregation. Federated learning overcomes these restrictions by decoupling the model training procedure from centralized data processing, ensuring private data is not exchanged. Figure~\ref{fig:federated} depicts a standard FL setup where a network of clients, which can vary in size from mobile devices to large computing clusters, collaboratively train a global model under the coordination of a central server. In each communication round, the server broadcasts the global model parameters to clients, who then update the model independently using their respective local datasets. These updates are subsequently sent back to the server, where they are aggregated to generate a new version of the global model. The original implementation of FL, known as FedAvg \cite{mcmahan_communication-efficient_2017}, uses averaging as the aggregation method. 

Although federated learning has gained significant traction due to its privacy-preserving capabilities, its efficacy predominantly depends on the assumption that all clients have homogeneous data distributions. However, in real-world applications, a significant challenge is \textbf{statistical heterogeneity}, where client data is not independent and identically distributed (non-IID) or balanced with respect to the number of data samples. This inherent diversity among local data distributions can lead to performance degradation as it becomes difficult to learn a global model that generalizes well to each client. Another major issue is \textbf{systems heterogeneity}, which refers to differences in clients' resources, such as communication bandwidth, storage capabilities, computational capacity, etc.

Efforts to augment FL algorithms with the ability to better manage heterogeneity are generally divided into two main strategies, based on the number of models trained. The first strategy aims to modify the global consensus model for greater robustness against non-IID datasets, typically using some form of model regularization. For instance, FedProx \cite{li2020federatedb} introduces a regularization term to the loss function, penalizing updates that significantly deviate from the global model. In recent years, personalized federated learning (pFL) has emerged as a more effective strategy, adopting a client-centric approach with the goal of training multiple personalized models. The key to pFL is finding the appropriate balance between utilizing client-specific knowledge derived from the local data and leveraging the shared knowledge among clients captured by the global model. While pFL systematically addresses the influence of statistical heterogeneity, we observe three critical limitations of existing methods: 

\begin{itemize}
    \setlength\itemsep{0.1em}
    \item \textbf{Homogeneous models} - Most of these methods depend on gradient-based aggregation, which is only viable if all clients share the same model architecture, a condition that is highly restrictive for various reasons. It may be infeasible for clients with fewer computational resources or hardware constraints to train a highly parameterized neural network. Additionally, clients may wish to use certain model architectures to meet specific requirements or individual preferences \cite{ye2023heterogeneous}. Moreover, these methods are unsuitable for use with non-differentiable models. 

    \item \textbf{Reliance on a central server} - Most efforts to advance pFL have focused on the centralized setting. In centralized federated learning (CFL), large service providers, such as organizations or research institutions, often serve as the central server. However, the costly overhead and maintenance of a central server may not be feasible for participating clients \cite{yuan2023decentralized}. As the number of clients grows, the central server can also become a communication bottleneck due to bandwidth limitations, network congestion, and the complexity of synchronizing updates from clients with diverse hardware capabilities \cite{beltran2023decentralized}. Furthermore, the existence of a central server introduces a single point of failure, making the network vulnerable to attacks that can compromise the privacy of all of the network’s constituents. 

    \item \textbf{Negative transfer} - It has been observed that personalized federated learning methods can be susceptible to negative transfer, where the performance for certain clients deteriorates due to collaboration with other clients \cite{rudovic2021personalized}. There has been minimal research exploring the factors that influence whether a client's participation in collaborative learning is advantageous or potentially detrimental to a client's local performance. To avoid negative transfer, an ideal strategy would incorporate a mechanism that defaults to relying solely on local information in scenarios where collaborative participation proves to be counterproductive.
\end{itemize}
 
In this work, we propose a novel algorithm, Federated Peer-Adaptive Ensemble Learning (FedPAE), which advances the practical applicability of pFL by supporting both systems and statistical heterogeneity. The basis of this learning framework is peer-to-peer model sharing, effectively eliminating the need for a central server. As models are shared within the network, clients benefit by learning from the diverse data distributions of their peers. This transition to complete decentralization is key for facilitating asynchronous learning, allowing clients to contribute and update models at their convenience. After exchanging models, every client in the network holds a repository of models (or model predictions when storage is limited) from which they independently construct an ensemble optimized for performance on their local data distribution. In cases where integrating external models could be counterproductive due to significant differences in data distributions, FedPAE provides clients the autonomy to rely exclusively on their locally trained models. This feature safeguards against the decrease in performance that would result from learning irrelevant information from other clients. An additional benefit of FedPAE is how it fine-tunes the balance between local data insights and the collective knowledge obtained from the network. While most pFL methods aim to balance contributions from local and global information, FedPAE offers a nuanced approach. Through ensemble selection, a client may incorporate models from peers if they enhance the local ensemble’s performance. Furthermore, a prominent feature of FedPAE is its capability to support model heterogeneity. By removing constraints on the model architecture, a client with resource constraints, such as a mobile device, can select lightweight models, whereas a more powerful server could opt for more complex, computationally expensive models. In summary, our key contributions are:

\begin{itemize}
\setlength\itemsep{0.1em}
\item \textbf{Development of a fully decentralized pFL framework}: We propose FedPAE, a novel algorithm that implements direct communication between clients via peer-to-peer model sharing to enable asynchronous learning.
\item \textbf{Support for model heterogeneity}: Our algorithm removes any constraints on model architecture, allowing clients to deploy models optimally suited to their local computational environment and specific learning tasks.
\item \textbf{Robust protection against statistical heterogeneity}: FedPAE enables clients to rely exclusively on local information in scenarios where contributions from other clients would be detrimental to the learning process.
\end{itemize}

This work is organized as follows: Section 2 reviews related work on ensemble selection methods and personalized federated learning. In Section 3, we introduce the proposed FedPAE framework and the optimization techniques used for ensemble selection. Experimental results are presented in Section 4, followed by a discussion in Section 5. Section 6 outlines the limitations of this study, and final conclusions are drawn in Section 7.

\section{Related Works}
\subsection{Federated Learning with non-IID data}

Due to the inherent benefit of circumventing the transmission of local data, FL methods are increasingly being adopted for learning tasks in domains with privacy-sensitive data. However, the presence of statistical and systems heterogeneity among clients poses constraints on the performance and practicality of a shared global model. Evidence demonstrates that classical federated learning methods such as FedAvg can encounter significant performance degradation when applied in non-IID settings \cite{karimireddy2020scaffold, li2020federated}.

Several works have been proposed to alleviate the negative impacts of heterogeneity experienced by the vanilla FedAvg algorithm. Model regularization is a minor modification to FedAvg that accounts for the dissimilarity between local objectives, allowing more robust and stable convergence. FedProx \cite{li2020federatedb} adds a proximal term to the local optimization problem of each client to mitigate issues arising from client drift, a phenomenon where heterogeneity causes significant divergence between the global and local models.
\citet{li2021fedbn} address feature shift (e.g., variations in medical images from different machines or protocols) by introducing FedBN, wherein local models have batch normalization layers that are exempt from parameter averaging.

SCAFFOLD \cite{karimireddy2020scaffold} estimates the update directions for both global and local models, using the difference between them to approximate client drift and adjust local updates accordingly. While early-stage efforts to address the limitations of the original FedAvg may offer some performance improvements in relatively mild non-IID scenarios, they still assume a “one-size-fits-all model” which may not be sufficient when there are significant differences among data distributions.   

\subsection{Personalized Federated Learning}

Personalized federated learning (pFL) has gained traction in recent years as a natural way to address the deficiencies of a single global model setup in more severe non-IID scenarios. A straightforward tactic for pFL is \textbf{local fine-tuning}, where a shared global model is initially trained and then customized for each client through a series of local gradient descent steps. Some recent works have drawn parallels between local fine-tuning and meta-learning by viewing each client as a learning task and employing a shared meta-learner in place of the conventional global model \cite{fallah2020personalized}. 

Within the domain of \textbf{regularization-based} techniques, DITTO \cite{li2021ditto} formulates pFL as a multi-task learning problem and introduces a regularization term to constrain the distance between the global and local models. Similarly, \citet{t2020personalized} propose pFedMe, which frames FL as a bi-level optimization problem that uses Moreau envelopes as clients’ regularized loss function to decouple the process of optimizing the global and local models. 

Among \textbf{layer personalization} methods, FedPer \cite{arivazhagan2019federated} and FedRep \cite{collins2021exploiting} split the model backbone into base and personalization layers to learn a shared global feature representation across clients while training personalization layers exclusively on local data. \textbf{Model interpolation} approaches learn personalized models using a mixture of the global and local models. For instance, APFL \cite{deng2020adaptive} uses a client-specific mixing parameter to control the level of contribution from the global model to each client's learning process.

Methods based on \textbf{personalized aggregation} focus on selectively collaborating with clients with similar learning objectives. FedFomo \cite{zhang2020personalized} estimates client-specific weighted combinations of the personalized models from other clients using a local validation set. Similar to FedFomo, APPLE \cite{luo2022adapt} determines a unique weighted combination of models for each client with the point of difference being that the weights are learned, rather than approximated. FedAMP \cite{huang2021personalized} is based on a message-passing mechanism that adaptively facilitates pairwise collaboration between clients with similar data distributions. 

\textbf{Clustering-based} methods aim to find a natural grouping of clients with similar data distributions, enabling the learning of personalized models within each cluster. \citet{sattler2020clustered} implement a bi-partitioning algorithm based on cosine similarity of gradient updates to group clients into clusters with jointly trainable distributions. \citet{ghosh2020efficient} propose an iterative clustering algorithm that alternates between estimating the cluster identities of clients and minimizing the loss functions within the clusters.

Integrating concepts from \textbf{decentralized federated learning}, several pFL methods focus on peer-to-peer communication to improve personalization by promoting collaboration among clients with similar data distributions. For instance, \citet{dai2022dispfl} introduce a decentralized sparse training technique (Dis-PFL) where each client maintains a fixed number of active parameters during the peer-to-peer communication process, ultimately reducing the communication bottleneck and saving local training costs. \citet{onoszko2021decentralized} propose Performance-Based Neighbor Selection (PENS), a gossip learning approach where clients with similar data distributions have a higher probability of collaborating.

\begin{figure*}[!b]
  \centering
  \includegraphics[height=7cm]{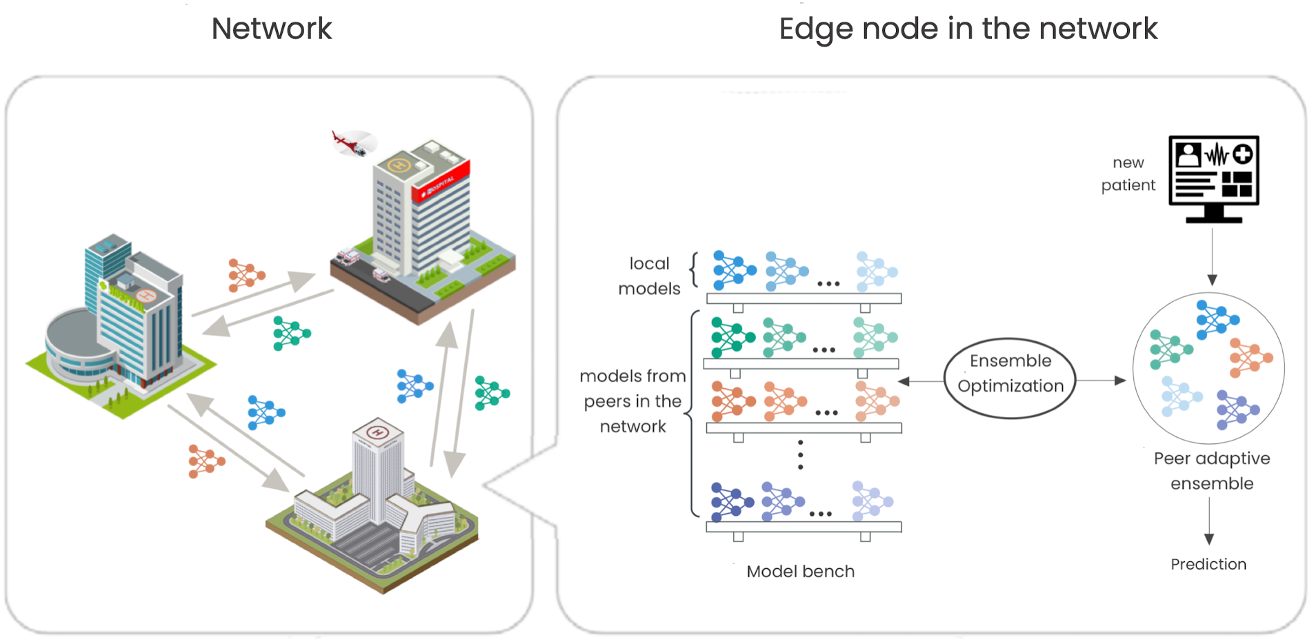}
\caption{Overview of Peer-Adaptive Ensemble Learning}
\label{fig:peer_adaptive}
\end{figure*}

\subsection{Model-heterogeneous federated learning}
One approach for model-heterogeneous FL is knowledge distillation (KD), where the goal is to transfer knowledge from a large, complex ‘teacher’ model to a more compact and simpler ‘student’ model. In the most basic variant of KD, the student model is trained to imitate the teacher model’s outputs on a proxy dataset by minimizing the discrepancy between the logits produced by the teacher and student model. In early implementations of KD in FL, such as FedMD \cite{li2019fedmd}, clients transmit predicted logits on a shared public dataset for aggregation at the server, which are subsequently redistributed to clients as global logits. However, the public datasets necessary for such approaches are not always readily available in practice. Moreover, the public datasets must be closely related to the learning task to achieve satisfactory model performance, which adds to the difficulty of sourcing appropriate data. 

Approaches like FML \cite{shen2020federated} and FedKD \cite{wu2022communication} train a smaller auxiliary model through mutual distillation, avoiding the need for a global dataset. Yet, in initial stages with limited feature extraction capabilities, the client and auxiliary models may negatively influence each other \cite{li2023smkd}. Some methods divide each client's local model into two segments: a feature extractor and a classifier header. During model aggregation, only one part is shared, while the other, containing unique parameters or diverse structures, is retained on the client's side. For example, LG-FedAvg \cite{liang2020think} and FedGH \cite{yi2023fedgh} share homogeneous classifier headers while maintaining heterogeneous feature extractors locally. A major drawback of the aforementioned model-heterogeneous approaches is their dependency on a centralized framework, where the central server is essential for aggregating and distributing information.

\subsection{Ensemble Selection}
Ensemble learning is a machine learning technique that combines the outputs of multiple classifiers to make a single prediction. This concept is analogous to a committee of experts, where a collective decision is generally more reliable than the perspective of a single individual. Ensemble methods work by mitigating the instability and random errors of individual classifiers, thereby enhancing model generalization \cite{webb2004multistrategy, zhang2012ensemble}. Several large-scale comparative studies on classification algorithms have demonstrated ensemble learning methods are able to achieve the best average classification accuracy on datasets that cover a wide spectrum of applications and domains \cite{10.5555/2627435.2697065, ZHANG2017128}. 
The literature provides evidence that an effective ensemble is one where the classifiers are both accurate and independent with respect to the errors they make \cite{hansen_neural_1990,krogh_neural_1994, opitz_generating_1995}. Ensemble selection methods, also known as ensemble pruning methods, aim to identify an accurate and diverse subset of classifiers that outperforms the original ensemble comprised of all available classifiers. For most problems, it's infeasible to search the entire solution space of all possible subsets. Numerous heuristic methods have been proposed to obtain reasonably good solutions more efficiently \cite{10.5555/645526.757762, zhou_ensembling_2002, CAVALCANTI201638, zhang2006ensemble}.

\section{Methods}

\subsection{Peer-Adaptive Ensemble Learning}

In the proposed learning framework, a network is comprised of multiple clients (e.g., hospitals, mobile devices, sensors). Each client utilizes its local data to train one or more models independently. After local training, clients engage in a peer-to-peer model exchange, where each client shares its models with other clients in the network. Differing from pFL methods that require clients to train a common model architecture, our model-heterogeneous algorithm allows clients the flexibility to employ different base classifiers. Consequently, every client holds a 'model bench' – a diverse collection of models acquired from peers as well as models trained locally. 

However, for clients with limited storage capacity, maintaining a large model bench may not be practical. To address concerns about memory and storage overhead, we propose a solution where clients can opt not to store models. Instead, they process the received models from peers to extract the predictions for local validation, allowing them to evaluate external models for ensemble selection without retaining them long-term. In this case, the ‘model bench’ consists of stored predictions, serving as compact representations of the models, which significantly reduces memory and storage demands. Once the optimal ensemble is selected, the client can download only the necessary models required for inference.

Peer-adaptive ensemble learning can be conceptualized as a collection of optimization problems, each defined by a unique data distribution. From the model bench, each client selects a subset of models to construct an ensemble optimized for performance on the local data distribution. Figure~\ref{fig:peer_adaptive} illustrates model sharing at the network level and ensemble selection at an individual client.

A key advantage of FedPAE is its robust protection against heterogeneity. In settings where the data distributions of clients are substantially different from each other, collaborative learning may be harmful to the local model performance. Through ensemble optimization, clients can exclude contributions from other clients' data by selecting an ensemble comprised solely of locally trained models. Some other pFL methods incorporate a regularization parameter to determine the global vs. local contribution to a client’s personalized models. By setting the regularization parameter to a certain value, these methods could essentially reduce the problem to individual client learning. However, these regularization terms are hyper-parameters that need to be determined before training. FedPAE offers an intrinsic way to reduce the problem to local learning.

\subsubsection{Model Selection and Ensemble Evaluation}

In the domain of ensemble learning, it is generally known that strength and diversity are central factors for ensemble performance \cite{dietterich2002ensemble}. Strength-based measures correspond to the performance of individual classifiers, which can be evaluated with metrics such as accuracy or AUC. While there is no single definition of diversity, the concept generally refers to the independence of classifiers in the ensemble. Research has shown that generalization error can be reduced when errors made by classifiers are uncorrelated or anti-correlated \cite{hansen_neural_1990,krogh_neural_1994,maclin1995combining, opitz_generating_1995}.

To perform ensemble selection, we implement a multi-objective optimization algorithm to simultaneously promote ensemble strength and diversity. Specifically, we use the non-dominated sorting genetic algorithm (NSGA-II), a popular evolutionary-based method that efficiently handles multiple conflicting objectives \cite{deb2002fast}. For ensemble strength, we consider the average accuracy among ensemble members. For ensemble diversity, we implement the same diversity measure as \citet{pang2019improving}, which measures the degree of independence between the predicted probabilities of a pair of classifiers. The core idea of NSGA-II is to determine the Pareto front that corresponds to a collection of non-dominated “Pareto-optimal” solutions. A solution is considered to be non-dominated if none of the objectives can be improved without simultaneously degrading another objective. The algorithm starts with a randomly initialized population of candidate solutions, each assigned a fitness score for both objectives. Each individual chromosome in the population encodes an ensemble as a binary string, where a 1 indicates a model is included in the ensemble and a 0 indicates model exclusion. In each generation, individuals are selected based on a dominance ranking scheme. 

\begin{figure}[t]
\centering
\includegraphics[width=0.45\textwidth]{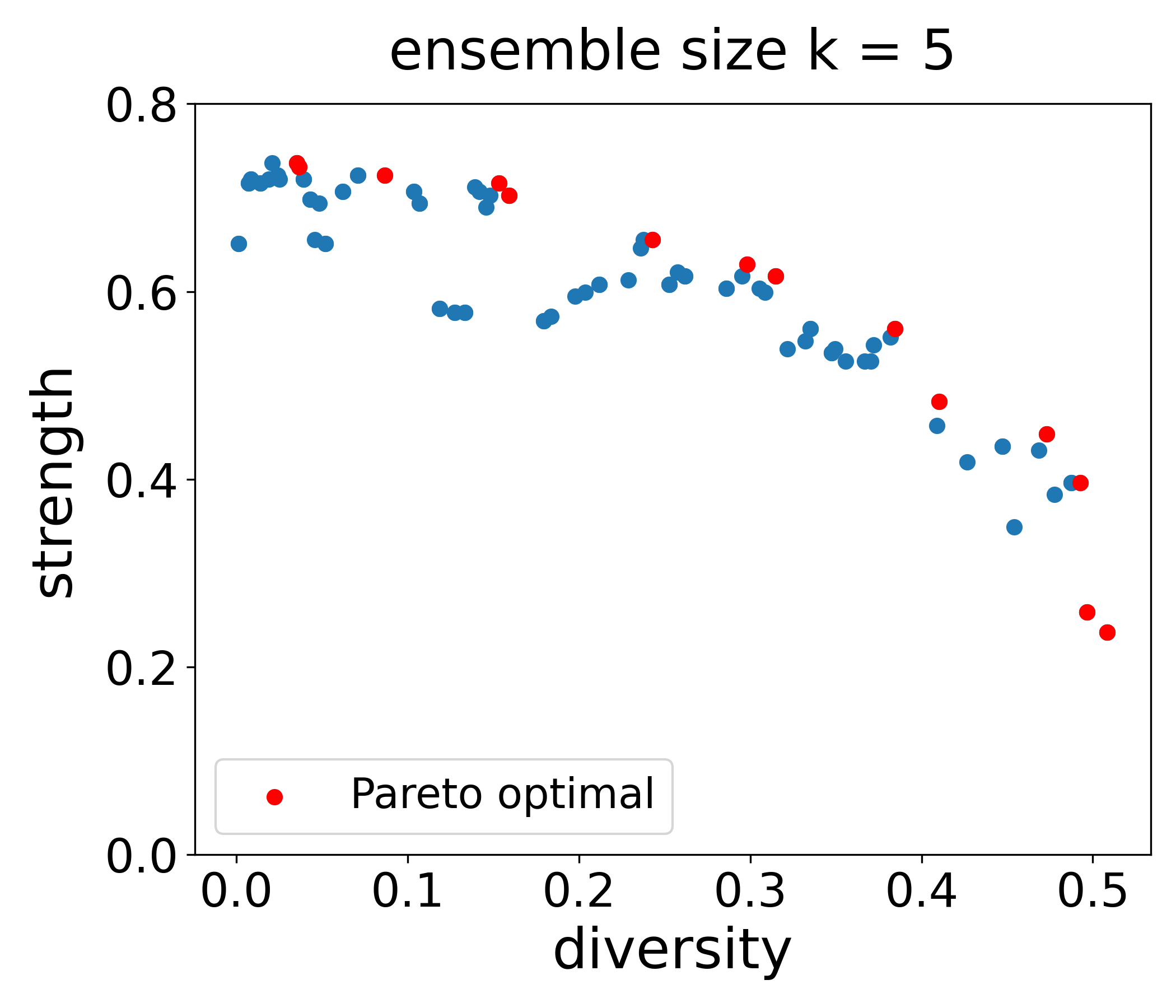}
\caption{Example Pareto front for a single client generated from the final generation of NSGA-II. Each point represents a potential ensemble, with the axes showing the trade-off between ensemble strength and diversity.}
\label{fig:pareto}
\end{figure}

Our motivation for applying NSGA-II lies in its ability to address the overfitting issues we encountered with previous iterations of ensemble selection algorithms, where the selected ensembles exhibited high accuracy and diversity on the validation set but failed to maintain high performance on the testing set. The filtering mechanism provided by NSGA-II allows us to refine the candidate set and focus on strong potential ensembles, while reducing the risk of selecting an ensemble that is overly tailored for performance on the validation set. 

The final generation’s output forms a Pareto front, as shown in Figure~\ref{fig:pareto}. We then evaluate the Pareto-optimal solutions and select the ensemble that achieves the highest overall accuracy on the validation set. It's important to note that the ensemble’s overall accuracy differs from the strength measure used as one of the objectives in NSGA. While the strength objective focuses on the average accuracy of individual classifiers, the overall accuracy reflects the collective performance of the ensemble.

\subsection{Experimental Design}
\begin{figure*}[t]
  \centering
  \includegraphics[height=5cm]{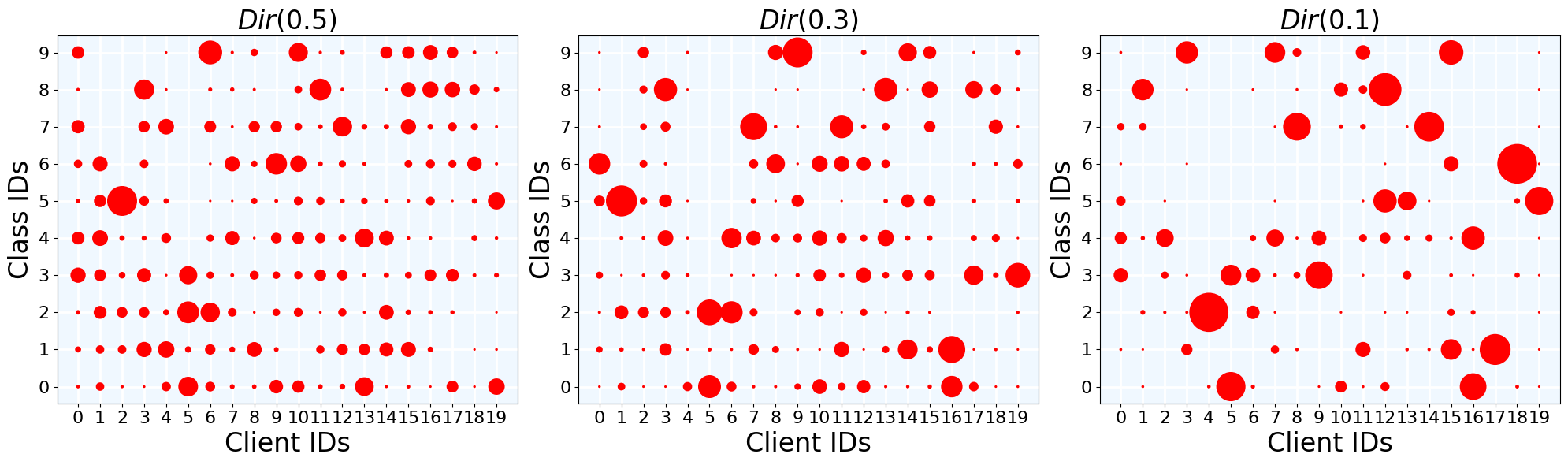}
\caption{Client data distributions on CIFAR-10 at three levels of statistical heterogeneity. The x-axis indicates client IDs, the y-axis indicates class labels, and the size of each point represents the number of data samples. As the Dirichlet distribution parameter decreases from 0.5 (left) to 0.1 (right), statistical heterogeneity increases. Higher heterogeneity results in clients having more samples concentrated in fewer class labels.}
\label{fig:dist}
\end{figure*}

\textbf{Datasets and Data Partition}. We evaluate FedPAE using two widely-used image classification datasets: CIFAR-10, with 60,000 images across 10 classes, and CIFAR-100, with 60,000 images across 100 classes \cite{krizhevsky2009learning}. To simulate non-IID settings, we partition the data for 20 clients following the Dirichlet distribution \cite{hsu2019measuring}, denoted as $Dir(\alpha)$. The parameter $\alpha$ controls the degree of heterogeneity, with smaller values corresponding to a greater data size imbalance and more diverse data distributions among clients. We observe the performance at three different levels of heterogeneity with $\alpha \in \{0.1, 0.3, 0.5\}$. It is important to note that the types of statistical heterogeneity can be categorized based on differences in the distribution of features $P(x)$, class priors $P(y)$, and conditional distributions $P(y|x)$. Although real-world applications often exhibit varying degrees of all types of data heterogeneity, we created diversity in our data samples by explicitly varying only $P(y)$, although it can be expected that $P(x)$ and $P(y|x)$ would also change as a result. Using the CIFAR-10 dataset as a reference, Figure~\ref{fig:dist} provides a visual representation of client data distributions across the different non-IID settings. Each client's local dataset is split into 70\% training, 15\% validation, and 15\% testing.

\begin{table*}[!b]
\caption{Mean test accuracy across clients with 95\% confidence intervals}
\label{tab:tab1}
\resizebox{\textwidth}{!}{
\begin{tabular}{l|ccc|ccc}
\hline
Dataset &
  \multicolumn{3}{c|}{Cifar10} &
  \multicolumn{3}{c}{Cifar100} \\ \hline
Heterogeneity &
  $Dir(0.5)$ &
  $Dir(0.3)$ &
  $Dir(0.1)$ &
  $Dir(0.5)$ &
  $Dir(0.3)$ &
  $Dir(0.1)$ \\ \hline
\cellcolor[HTML]{C0C0C0}\textbf{FedAvg} &
  \cellcolor[HTML]{FFFFFF}0.697 ± 0.025 &
  \cellcolor[HTML]{FFFFFF}0.677 ± 0.028 &
  \cellcolor[HTML]{FFFFFF}0.668 ± 0.062 &
  \cellcolor[HTML]{FFFFFF}0.352 ± 0.012 &
  \cellcolor[HTML]{FFFFFF}0.347 ± 0.019 &
  \cellcolor[HTML]{FFFFFF}0.332 ± 0.017 \\
\cellcolor[HTML]{C0C0C0}\textbf{FedProx} &
  \cellcolor[HTML]{FFFFFF}0.697 ± 0.025 &
  \cellcolor[HTML]{FFFFFF}0.675 ± 0.029 &
  \cellcolor[HTML]{FFFFFF}0.667 ± 0.062 &
  \cellcolor[HTML]{FFFFFF}0.351 ± 0.011 &
  \cellcolor[HTML]{FFFFFF}0.348 ± 0.018 &
  \cellcolor[HTML]{FFFFFF}0.330 ± 0.018 \\ \hline
\cellcolor[HTML]{C0C0C0}\textbf{FedDistill} &
  \cellcolor[HTML]{FFFFFF}0.730 ± 0.039 &
  \cellcolor[HTML]{FFFFFF}0.799 ± 0.036 &
  \cellcolor[HTML]{FFFFFF}0.867 ± 0.052 &
  \cellcolor[HTML]{FFFFFF}0.342 ± 0.029 &
  \cellcolor[HTML]{FFFFFF}0.415 ± 0.033 &
  \cellcolor[HTML]{FFFFFF}0.555 ± 0.030	 \\
\cellcolor[HTML]{C0C0C0}\textbf{LG-FedAvg} &
  \cellcolor[HTML]{FFFFFF}0.733 ± 0.039 &
  \cellcolor[HTML]{FFFFFF}0.803 ± 0.037 &
  \cellcolor[HTML]{FFFFFF}0.872 ± 0.048 &
  \cellcolor[HTML]{FFFFFF}0.313 ± 0.025 &
  \cellcolor[HTML]{FFFFFF}0.412 ± 0.032 &
  \cellcolor[HTML]{FFFFFF}0.523 ± 0.026	 \\
\cellcolor[HTML]{C0C0C0}\textbf{FedKD} &
  \cellcolor[HTML]{FFFFFF}0.729 ± 0.038 &
  \cellcolor[HTML]{FFFFFF}0.806 ± 0.034 &
  \cellcolor[HTML]{FFFFFF}0.870 ± 0.048 &
  \cellcolor[HTML]{FFFFFF}0.329 ± 0.026 &
  \cellcolor[HTML]{FFFFFF}0.426 ± 0.028 &
  \cellcolor[HTML]{FFFFFF}0.539 ± 0.028 \\
\cellcolor[HTML]{C0C0C0}\textbf{FedGH} &
  \cellcolor[HTML]{FFFFFF}0.731 ± 0.040 &
  \cellcolor[HTML]{FFFFFF}0.792 ± 0.038 &
  \cellcolor[HTML]{FFFFFF}0.869 ± 0.052 &
  \cellcolor[HTML]{FFFFFF}0.329 ± 0.032 &
  \cellcolor[HTML]{FFFFFF}0.406 ± 0.034 &
  \cellcolor[HTML]{FFFFFF}0.541 ± 0.032	 \\
\cellcolor[HTML]{C0C0C0}\textbf{FML} &
  \cellcolor[HTML]{FFFFFF}0.729 ± 0.040 &
  \cellcolor[HTML]{FFFFFF}0.796 ± 0.038 &
  \cellcolor[HTML]{FFFFFF}0.870 ± 0.049 &
  \cellcolor[HTML]{FFFFFF}0.320 ± 0.029 &
  \cellcolor[HTML]{FFFFFF}0.394 ± 0.030 &
  \cellcolor[HTML]{FFFFFF}0.528 ± 0.027	 \\
\cellcolor[HTML]{C0C0C0}\textbf{local} &
  \cellcolor[HTML]{FFFFFF}0.744 ± 0.029 &
  \cellcolor[HTML]{FFFFFF}0.805 ± 0.033 &
  \cellcolor[HTML]{FFFFFF}0.871 ± 0.046 &
  \cellcolor[HTML]{FFFFFF}0.381 ± 0.019 &
  \cellcolor[HTML]{FFFFFF}0.445 ± 0.022 &
  \cellcolor[HTML]{FFFFFF}0.556 ± 0.020 \\ \hline
\cellcolor[HTML]{C0C0C0}\textbf{FedPAE} &
  \cellcolor[HTML]{FFFFFF}\textbf{0.774 ± 0.026} &
  \cellcolor[HTML]{FFFFFF}\textbf{0.826 ± 0.030} &
  \cellcolor[HTML]{FFFFFF}\textbf{0.873 ± 0.047} &
  \cellcolor[HTML]{FFFFFF}\textbf{0.399 ± 0.016} &
  \cellcolor[HTML]{FFFFFF}\textbf{0.457 ± 0.020} &
  \cellcolor[HTML]{FFFFFF}\textbf{0.558 ± 0.020}
\end{tabular}
}
\end{table*}

\textbf{Baselines.} 
We compare FedPAE with a local ensemble baseline, 2 traditional federated learning methods, and 5 state-of-the-art personalized federated learning methods that can accommodate heterogeneous model architectures. The local ensemble baseline refers to the collection of models that each client trains independently on their local data. For the traditional FL baselines methods, we use the classic FedAvg algorithm, and the regularization-based adaptation, FedProx. The pFL methods include FedKD, FML, FedGH, LG-FedAvg, and FedDistill.

\textbf{Models.} For model-heterogeneous settings, we use 5 different model architectures including the 4-layer CNN adopted from \cite{mcmahan_communication-efficient_2017}, ResNet-18 \cite{he2016deep}, Densenet-121 \cite{huang2017densely}, GoogleNet \cite{szegedy2015going}, and VGG-11 \cite{simonyan2014very}. In FedPAE, each client undergoes training for all five models, which are then shared with every other client in the network. For the pFL baselines, the five model architectures are distributed evenly across clients, with the first client assigned a CNN, the second a ResNet-18, the third a DenseNet-121, and so on. Since LG-FedAvg and FedGH rely on homogeneous classifier headers, we consider the last FC layer as homogeneous for parameter aggregation on the server, while the remaining layers have heterogeneous structures. Following the principles of FedKD and FML, it is advisable to keep the auxiliary model as compact as possible to minimize communication overhead during the transmission of model parameters. Therefore, we select the smallest model architecture model, the 4-layer CNN, to serve as the auxiliary model for both FedKD and FML. In line with this reasoning, the traditional FL methods also adopt CNN as the common model architecture shared among clients. 

\begin{figure*}[!t]
  \centering
  \includegraphics[height=5cm]{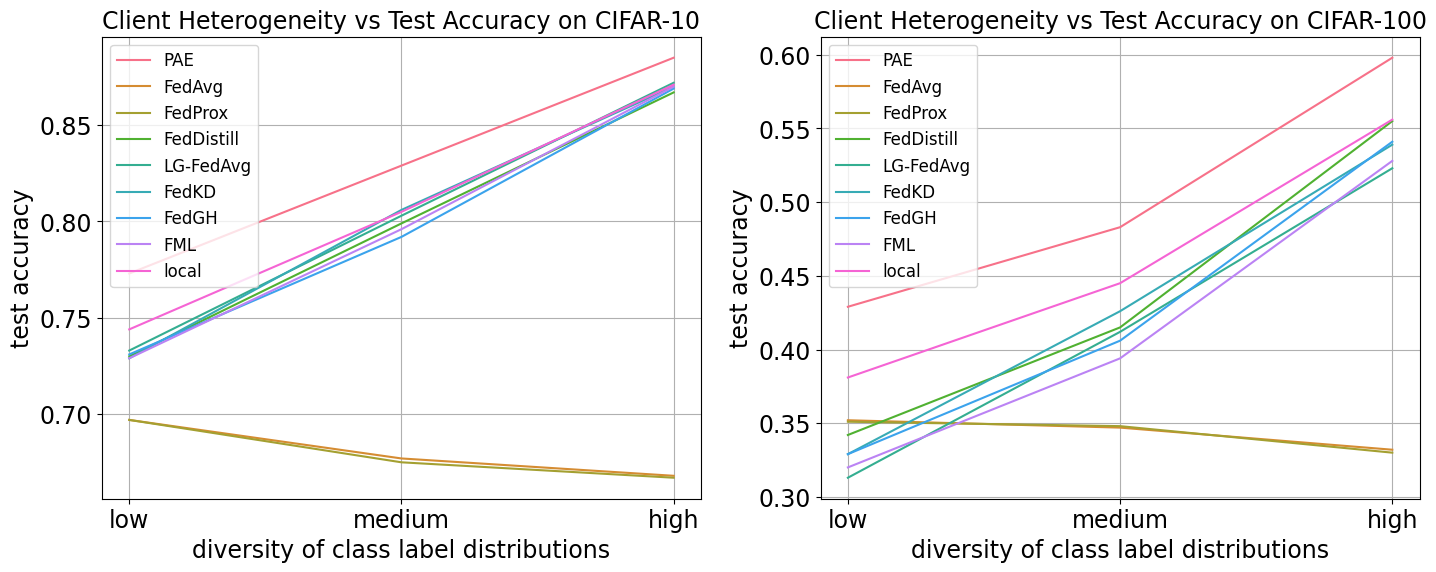}
\caption{Performance on CIFAR-10 and CIFAR-100 across different levels of heterogeneity}
\label{fig:output}
\end{figure*}

\textbf{Implementation details.} The models for FedPAE were trained for up to 500 epochs, incorporating an early stopping mechanism that halted training if no improvement was observed after 50 consecutive epochs. For comparison methods, the training process spanned 500 communication rounds with one local step performed in each round. We adopted a consistent learning rate of 0.01 and set the mini-batch size to 10. For FedPAE and any baselines with special hyper-parameters, we performed a grid search to determine the optimal settings. Specifically, for the NSGA-II algorithm used in ensemble selection, we selected the key hyper-parameters, population size and number of generations, by evaluating the average accuracy on the validation set across different configurations. The number of generations was varied between 100 and 400, and the population size between 50 and 200, with the final settings of 100 generations and a population size of 100 chosen based on the best validation performance. For FedPAE, each client constructed an ensemble of size $k = 5$. For all methods, we monitored training using the validation set, and the final model state chosen for evaluation corresponded to the point where validation accuracy was maximized. It's important to note that the other baseline methods do not perform this step in their original implementation. Rather, they assume all clients will utilize the personalized model obtained after the final epoch of training. The use of a validation set to monitor progress during training and determine the performance optimized for each client provides an extra degree of personalization. For FedPAE, the validation set serves the dual purpose of determining the early stopping point during training and guiding classifier selection during ensemble optimization. All experiments were conducted on a high-performance computing cluster, utilizing nodes with 64-core CPUs, 384 GB of RAM, and four NVIDIA GeForce RTX 2080 Ti GPUs.

\section{Results}

\textbf{Heterogeneity.}
Table \ref{tab:tab1} summarizes our results. We report the unweighted average accuracy on the clients' test sets with standard deviations. In the context of CIFAR-10, FedPAE consistently outperformed traditional FL methods across all heterogeneity settings. Notably, under the highest degree of heterogeneity, FedPAE achieved an accuracy of 0.873, surpassing the performance of FedAvg and FedProx by 32.5\% and 32.7\%, respectively. As the heterogeneity decreased, the performance gaps between FedPAE and FedAvg/FedProx narrowed, yet FedPAE still showed a substantial advantage. Figure~\ref{fig:output} illustrates the performance trends as the diversity in client data distributions increases.

In comparison to the pFL baselines, FedPAE also demonstrated superior performance across all levels of heterogeneity. Following the trend of FedPAE, the performance of the other pFL methods improves as the degree of heterogeneity increases. With a fixed number of data samples per client, increasing heterogeneity leads to increased entropy within the class distributions, which may partially contribute to the improvement in performance. With fewer class labels to learn, the more effective personalization becomes because clients can specialize in the distinct patterns of their local data. 

In the experiments using the CIFAR-100 dataset, FedPAE also exhibited leading performance with accuracy gains over FedAvg and FedProx being even more pronounced due to the increased task complexity. Under severe heterogeneity, FedPAE achieved an accuracy of 0.558, which also surpasses the performance of the pFL baselines. At the lowest level of heterogeneity for CIFAR0-100, we also observe that FedProx and FedAvg start to outperform the other pFL baselines.  

As heterogeneity increased, we also observed variation in the average percentage of locally trained models selected by clients for their ensemble. In the CIFAR-100 experiments, for instance, clients selected locally trained models 55\% of the time under $Dir(0.5)$, 66\% under $Dir(0.3)$, and 72\% under $Dir(0.1)$. These findings underscore the increased reliance on local models in highly heterogeneous settings, where the ensemble selection process increasingly favors local models as integrating external models becomes less effective or even detrimental.

To emphasize FedPAE's robustness to heterogeneity, we also examine the relative change in average test accuracy from the local ensemble baseline in the most heterogeneous setting for CIFAR-10. Table \ref{tab:tab2} shows that FedPAE nearly always enhances performance, with the most substantial decrease being -1.4\%,  while clients employing other pFL methods see performance declines ranging from -11.6\% to -7.0\%. In summary, FedPAE consistently preserves or improves client performance, making it a reliable safeguard against performance degradation in environments with substantial heterogeneity.

\begin{table}[h]
\centering
\caption{Range of the relative change in average test accuracy across clients from the local ensemble baseline under the highest level of statistical heterogeneity for CIFAR-10.}
\label{tab:tab2}
\begin{small}
\vskip 0.1in
\begin{tabular}{rc}
\cline{1-2}
\multicolumn{1}{l|}{Heterogeneity}                                                   & Dir(0.1)      \\ \hline
\multicolumn{1}{l|}{\cellcolor[HTML]{C0C0C0}{\color[HTML]{000000} \textbf{FedDistill}}}    & (-7.0\%, 11.9\%) \\
\multicolumn{1}{l|}{\cellcolor[HTML]{C0C0C0}{\color[HTML]{000000} \textbf{LG-FedAvg}}} & (-7.6\%, 9.4\%)   \\
\multicolumn{1}{l|}{\cellcolor[HTML]{C0C0C0}{\color[HTML]{000000} \textbf{FedKD}}}   &  (-9.6\%, 9.6\%)   \\
\multicolumn{1}{l|}{\cellcolor[HTML]{C0C0C0}{\color[HTML]{000000} \textbf{FedGH}}}  & (-11.3\%, 8.6\%)      \\
\multicolumn{1}{l|}{\cellcolor[HTML]{C0C0C0}{\color[HTML]{000000} \textbf{FML}}}   & (-11.6\%, 8.8\%) \\ \hline
\multicolumn{1}{l|}{\cellcolor[HTML]{C0C0C0}{\color[HTML]{000000} \textbf{FedPAE}}}      & (-1.4\%, 10.6\%)  
\end{tabular}
\vskip 0.1in
\end{small}
\end{table}

\textbf{Scalability.}
To demonstrate FedPAE's scalability, we perform additional experiments with 50 clients on the CIFAR-100 dataset in the most heterogeneous setting. With the total amount of data for CIFAR-100 remaining constant, the data available per client decreases as the number of clients increases, exacerbating the issue of insufficient local data. While FedKD slightly outperforms FedPAE in this setting, Table \ref{tab:tab3} shows that FedPAE still maintains very competitive performance as the network scales.

\begin{table}[h]
\centering
\caption{Average test accuracy with a network size of 50 clients under the highest level of statistical heterogeneity for CIFAR-100.}
\label{tab:tab3}
\begin{small}
\vskip 0.1in
\begin{tabular}{rc}
\hline
\multicolumn{1}{l|}{Heterogeneity}                                                   & Dir(0.1)      \\ \hline
\multicolumn{1}{l|}{\cellcolor[HTML]{C0C0C0}{\color[HTML]{000000} \textbf{FedAvg}}}  & 0.363 ± 0.017 \\
\multicolumn{1}{l|}{\cellcolor[HTML]{C0C0C0}{\color[HTML]{000000} \textbf{FedProx}}} & 0.364 ± 0.018 \\ \hline
\multicolumn{1}{l|}{\cellcolor[HTML]{C0C0C0}{\color[HTML]{000000} \textbf{FedDistill}}}    & 0.524 ± 0.032 \\
\multicolumn{1}{l|}{\cellcolor[HTML]{C0C0C0}{\color[HTML]{000000} \textbf{LG-FedAvg}}} & 0.535 ± 0.034 \\
\multicolumn{1}{l|}{\cellcolor[HTML]{C0C0C0}{\color[HTML]{000000} \textbf{FedKD}}}   & \textbf{0.554 ± 0.029} \\
\multicolumn{1}{l|}{\cellcolor[HTML]{C0C0C0}{\color[HTML]{000000} \textbf{FedGH}}}  & 0.537 ± 0.034 \\
\multicolumn{1}{l|}{\cellcolor[HTML]{C0C0C0}{\color[HTML]{000000} \textbf{FML}}}   & 0.518 ± 0.033 \\
\multicolumn{1}{l|}{\cellcolor[HTML]{C0C0C0}{\color[HTML]{000000} \textbf{local}}}   & 0.550 ± 0.027 \\ \hline
\multicolumn{1}{l|}{\cellcolor[HTML]{C0C0C0}{\color[HTML]{000000} \textbf{FedPAE}}}      & {0.552 ± 0.028}
\end{tabular}
\end{small}
\end{table}

\textbf{Computation cost.}

Table \ref{tab:computation} presents a comparison of computational complexity, focusing on the highest level of statistical heterogeneity examined in the CIFAR100 experiments, measured in terms of FLOPs (Floating Point Operations) and runtime at convergence (in minutes). FLOPs are computed by first obtaining the number of operations for each model architecture. Following the work of \citet{chiang2023mobiletl}, we assume that the backward pass during training requires twice the number of operations as the forward pass, so FLOPs are multiplied by 3 to reflect 1 training iteration. 

In the case of FedPAE, the computational complexity arises from training the base classifiers and ensemble selection. In the NSGA phase, each client selects an optimal ensemble of classifiers by evaluating a population  $P$  over  $G$  generations, contributing an additional complexity term of  $P \times G$. In the final step of ensemble selection, the Pareto optimal ensembles are evaluated on a validation set, with the cost of each evaluation being the FLOPs for a forward pass of each classifier. When considering all  $N$  clients and $M$ locally trained classifiers per client, the total computational complexity for FedPAE is $\mathcal{O}(N(M \times T \times D + P \times G + pf \times V))$, where $T$ is the number of training iterations for base classifiers, $D$ is the number of training samples, $V$ is the number of validation samples, and $pf$ is the average number of Pareto-optimal solutions that need to evaluated. In comparison to other methods which require multiple communication rounds, FedPAE exhibits a significant reduction in computational cost.

\begin{table}[h]
\centering
\caption{Comparison of Computational Complexity (in GFLOPs and runtime) under the highest level of statistical heterogeneity for CIFAR-100.}
\label{tab:computation}
\begin{small}
\begin{tabular}{lll}
\hline
\multicolumn{1}{c|}{\textbf{Method}}    & \multicolumn{1}{c|}{\textbf{GFLOPs}} & \multicolumn{1}{c}{\textbf{Runtime (mins)}} \\ \hline
\multicolumn{1}{l|}{\cellcolor[HTML]{C0C0C0}FedAvg}         & $6.33 \times 10^7$  & 192.38  \\
\multicolumn{1}{l|}{\cellcolor[HTML]{C0C0C0}FedProx}        & $6.33 \times 10^7$  & 208.37  \\ \hline
\multicolumn{1}{l|}{\cellcolor[HTML]{C0C0C0}LG-FedAvg}      & $3.56 \times 10^7$  & 860.43  \\
\multicolumn{1}{l|}{\cellcolor[HTML]{C0C0C0}FedKD}          & $1.88 \times 10^7$  & 498.65  \\
\multicolumn{1}{l|}{\cellcolor[HTML]{C0C0C0}FedGH}          & $4.54 \times 10^7$  & 1314.51 \\
\multicolumn{1}{l|}{\cellcolor[HTML]{C0C0C0}FedDistill}     & $3.44 \times 10^7$  & 1589.65 \\
\multicolumn{1}{l|}{\cellcolor[HTML]{C0C0C0}FML}            & $3.37 \times 10^7$  & 211.23  \\ \hline
\multicolumn{1}{l|}{\cellcolor[HTML]{C0C0C0}FedPAE}         & $1.27 \times 10^7$  & 108.92  \\ \hline
\end{tabular}
\end{small}
\end{table}

\section{Discussion}
The results obtained from our experiments support the efficacy of FedPAE in addressing the dual challenges of statistical and systems heterogeneity inherent in federated learning. FedPAE's superior performance across different levels of heterogeneity and classification tasks underscores its capability to handle non-IID data effectively. While other pFL methods showed varying degrees of success, FedPAE's approach to ensemble selection allows for a more nuanced balance between leveraging global knowledge and maintaining local relevance. This selective integration of models effectively combats negative transfer, where contributions from models trained on different data distributions may have adverse effects. FedPAE's model-heterogeneous nature allows for the inclusion of different model architectures, from lightweight to more advanced, catering to the computational constraints and personalized requirements of different clients. This flexibility, absent in traditional FL methods and many pFL algorithms is crucial for adapting to varying client capabilities. The elimination of the central server in FedPAE not only alleviates scalability concerns but also facilitates asynchronous learning. This allows clients to contribute to update the local ensemble at their convenience, without the communication bottleneck often encountered in the centralized setting.

\section{Limitations}
One limitation of FedPAE lies in its communication overhead. While other methods may have lower communication overhead compared to FedPAE, particularly in terms of the number of parameters transferred, this advantage is less significant in the broader context of runtime. Methods like FedKD, FML, and FedDistill leverage lightweight auxiliary models, logits, or learned local representations during communication, resulting in smaller data transfers. Despite these advantages in terms of total communication overhead, the overall runtime in our experiments would be primarily driven by computational complexity. The differences in communication time due to differences in data size is negligible compared to the substantial differences in runtime caused by computational complexity. Although FedPAE involves larger data transfers, its asynchronous nature allows clients to operate independently, mitigating the synchronization bottlenecks seen in other methods.

A potential strategy to reduce communication overhead in future experiments is to incorporate clustering-based approaches, where clients form smaller sub-networks. In FedPAE, for instance, clients could leverage historical data on model selection frequencies and prioritize collaboration with peers whose models are consistently selected during ensemble optimization. Additionally, clients could periodically reevaluate models from outside their cluster, allowing other clients the opportunity to establish themselves as potential collaborators in future ensemble selections.

\section{Conclusion}
In this work, we propose FedPAE, a decentralized and model-heterogeneous pFL method that leverages ensemble learning. Experiments performed with 8 baselines and 2 benchmark datasets demonstrate FedPAE's superior performance across different levels of statistical heterogeneity. The implications of FedPAE's performance are significant for real-world applications in domains with stringent privacy requirements and diverse client populations. FedPAE stands out as a robust solution for personalized federated learning, capable of handling the challenges posed by heterogeneous data distributions and system capabilities. 

In future work, we aim to enhance personalization through the implementation of dynamic ensemble selection. This advancement will shift the focus from optimizing ensemble performance for the general local data distribution to selecting a uniquely tailored ensemble for each individual test sample. With this modification, our approach would account for sample-specific variability. This feature is particularly attractive in domains such as healthcare, where it's desirable to accommodate the unique characteristics inherent in individual patient data.

\bibliographystyle{IEEEtranN}
\bibliography{bibliography}

\end{document}